\documentclass{article}
\usepackage{spconf,amsmath,graphicx}
\usepackage{framed,multirow}
\usepackage{amssymb}
\usepackage{caption}
\usepackage{multicol}
\usepackage{stfloats}
\usepackage{comment}
\usepackage{latexsym}
\usepackage{graphicx}
\usepackage{amsmath}
\usepackage{mathtools}
\usepackage{hyperref}
\usepackage{float}
\usepackage[font=small,labelfont=bf]{caption}
\usepackage{url}
\usepackage{cite}
\usepackage{xcolor}

\DeclareMathOperator*{\E}{\mathbb{E}}

\title{RSINET: INPAINTING REMOTELY SENSED IMAGES USING TRIPLE GAN FRAMEWORK}

\name{Advait Kumar$^1$$^\dagger$\sthanks{Equal contribution author} \qquad Dipesh Tamboli$^\ddag$\footnotemark[1] \qquad Shivam Pande$^2$$^\dagger$ \qquad Biplab Banerjee$^2$$^\dagger$}
\address{$^{1}$ Dept. of Electrical Engineering \qquad$^{2}$ Centre of Studies in Resource Engineering\\ $^\dagger$Indian Institute of Technology Bombay\\  \qquad$^\ddag$School of Electrical and Computer Engineering, Purdue University}

\begin{document}

\maketitle

\begin{abstract}
We tackle the problem of image inpainting in the remote sensing domain. Remote sensing images possess high resolution and geographical variations, that render the conventional inpainting methods less effective. This further entails the requirement of models with high complexity to sufficiently capture the spectral, spatial and textural nuances within an image, emerging from its high spatial variability. To this end, we propose a novel inpainting method that individually focuses on each aspect of an image such as edges, colour and texture using a task specific GAN. Moreover, each individual GAN also incorporates the attention mechanism that explicitly extracts the spectral and spatial features. To ensure consistent gradient flow, the model uses residual learning paradigm, thus simultaneously working with high and low level features. We evaluate our model, alongwith previous state of the art models, on the two well known remote sensing datasets, Open Cities AI and Earth on Canvas, and achieve competitive performance. The code can be referred here: \href{https://github.com/advaitkumar3107/RSINet}{\textcolor{cyan}{\underline{\textit{https://github.com/advaitkumar3107/RSINet}}}}.
\end{abstract}
\section{Introduction}
\label{sec:intro}

Image inpainting is the process of filling in the missing part or conserving the damaged and deteriorated image (which can be physical or digital). Digital image inpainting is an important problem statement in the field of computer vision which has applications in various domains such as restoring damaged images/videos, remote sensing, object removal, text removal, automatic modifications of images/videos, image compression and super resolution \cite{elharrouss2019image}. The drawback of the traditional mathematical methods is their low PSNR on complex images such as remote sensing images \cite{dong2018inpainting}. Remote sensing images have a variety of boundaries, objects and colours which makes it a semantically tougher problem. Hence, deep learning has been introduced earlier to tackle this challenge such as in \cite{dong2018inpainting}, which solved three missing information tasks in remote sensing data using a deep convolutional network combined with spatio-temporal information. Similarly, \cite{9116347} used a generative approach for image inpainting. These images looked quite similar to the original version. 

There have been several image inpainting approaches as well, such as \cite{liu2017deep}, which used a UNet based encoder-decoder structure. Furthermore, \cite{liu2017deep} also used residual learning to compute the gradients effectively for image restoration. \cite{yan2018shiftnet} built on the above idea and used guidance loss for inpainting. \cite{liu2018image} proposed a new type of layer called `partial convolution' (PartialConv) to improve the current best performing image inpainting models. However, all the above models used simple baselines, which resulted in lower PSNR on complex satellite images. \cite{nazeri2019edgeconnect} proposed the EdgeConnect (EC) model, that initially builds up the outline of the image to be restored and then fills it with details. Furthermore, to ensure the relationship between the neighbouring areas and entire image as a whole, \cite{10.1145/3072959.3073659} proposed a multi GAN approach, that trains a global as well as a local discriminator (GLCIC) on top of their VGG-type model \cite{simonyan2014deep}. Moreover, \cite{uddin20} proposed a global and local attention based model to tackle the issue of image irregularities such as holes. In addition, to preserve the semantic style of the original images, additional loss functions have been explored. For instance, \cite{Gatys2016ImageST} introduces style loss, where the textural features are synthesised from the images, which assist in the style transfer. Similarly, \cite{johnson2016perceptual} introduces perceptual loss for super-resolution, (instead of pixelwise loss) and get high resolution images. 

In the aforementioned methods, even though the methods seemed to perform better individually on the task specific images, they could not prove much effective on the images with high spatial variation such as those from remote sensing domain. This is because each of the individual models was designed to focus on a specific aspect of the image such as colour, edge or texture. Hence, there arises a need to have a common model that simultaneously works with all the image aspects and leads to more accurate image restoration. Inspired from this notion, we propose an image inpainting approach that utilizes multiple GANs to effectively capture the different aspects of the image. In addition, to handle irregularities of the image defects, Convolutional Block Attention Module (CBAM) \cite{woo2018cbam} and the Gated Attention Layers \cite{oktay2018attention} based attention modules are added. The GANs are also reinforced with skip connection to ensure a consistent gradient flow. Moreover, to preserve the semantic style of the images and get more robust representation, we incorporate a cocktail of adversarial, style and perceptual loss. Our approach can be summarised as follows:
\begin{enumerate}
    \item We introduce a model that combines the characteristics of edge detection GAN, colour filling GAN and an global GAN to focus on the different image aspects. 
    \item We introduce attention layers for guided backpropagation to get more robust spectral-spatial representation and increase the sharpness of the images, while simultaneously incorporating style and perceptual losses with the adversarial loss to capture the semantic information of the images.
    \item We train our model on the Open Cities AI \cite{naphade20192019} and the Earth on Canvas \cite{chaudhuri2021zero} datasets where our model outperforms the state of the art deep inpainting models by at least 0.06 \% and 2.24 \% respectively for the two datasets. 
\end{enumerate}

\section{Model Description}
\begin{figure*}
    \centering
    \includegraphics[width=17cm]{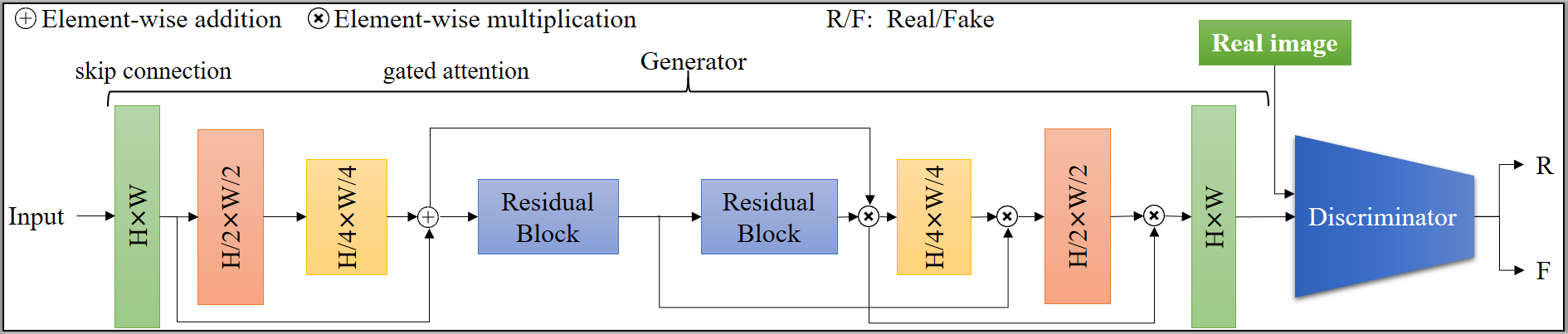}
    \caption{Architecture of our proposed GAN. The Generator consists of an encoder-residual blocks-decoder structure. The encoder has a skip connection as well as feature maps that serve as a gate for the gated attention layers. The Discriminator consists of 6 convolutional layers that successively downsample the input into a 28 $\times$ 28 probabilistic grid.}
    \label{fig:prop_arch}
\end{figure*}
In this section we discuss our proposed model. It consists of three GANs trained on top of each other, namely, the edge completion GAN $\mathcal{G}_1$, colour filling GAN $\mathcal{G}_2$ and the global GAN $\mathcal{G}_3$. We describe the loss functions and the architecture used for the basic GAN unit in detail in the subsequent sections. The GAN architecture used in RSINet is presented in Fig. \ref{fig:prop_arch}. All the GANs follow the similar architecture. 

Let $I_{gt}$ be the groundtruth image and, \textbf{$C_{gt}$} and \textbf{$I_{gray}$} be the groundtruth edge map (after applying a canny edge detector \cite{bao2005canny} to the true image) and grayscale image respectively. In the generator for $\mathcal{G}_1$, we use the masked grayscale image \textbf{$\tilde{I}_{gray} = I_{gray} \odot (1 - M)$}, its masked edge map \textbf{$\tilde{C}_{gt} = C_{gt} \odot (1 - M)$} ($\odot$ being the Hadamard product), and the image masks ${M}$ as the pre-condition (proposed in \cite{isola2016imagetoimage} as input to pix2pix GAN, where 1 and 0 denote the masked region and the background respectively). The generator produces the completed edge map for the image (see Eqn. \ref{eqn:ecomp}).
\begin{equation}
    C_{pred} = \mathcal{G}_1(\tilde{I}_{gray}, \tilde{C}_{gt}, M)
    \label{eqn:ecomp}
\end{equation}
We use $C_{gt}$ and $C_{pred}$ conditioned on $I_{gray}$ as the input to the discriminator, which predicts whether or not the edge map is real. This predicted edge map is then passed on to the colour filling model. The colour filling model uses the masked RGB image, $\tilde{I}_{gt} = I_{gt} \odot (1 - M)$, conditioned on the completed edge map $C_{comp} = C_{gt} \odot (1 - M) + C_{pred} \odot M$, as taken from the previous model. The model outputs the completed RGB image with the right colours filled at the right places. 
\begin{equation}
    I_{pred} = \mathcal{G}_2 \left(\tilde{I}_{gt}, C_{comp}\right)
\end{equation}
The previous two models gave blurry outputs on the tougher satellite images. Hence, we propose to train a global GAN on top of the output from the colour filling model which helps in refining the output and makes the image visually sharper. This was based upon the suggestion \cite{singh2018self} that using the same encoder-decoder architectures for image inpainting were observed to give better results since both models learnt the same features semantically. 
\begin{equation}
    I_{refined} = \mathcal{G}_3(I_{comp}, \tilde{I}_{gt})
    \label{eqn:glo}
\end{equation}

\subsection{Loss Functions}

The edge completion model is trained the adversarial loss \cite{isola2016imagetoimage} and the feature matching loss \cite{salimans2016improved}. Adversarial training is thought of as a min-max game between two players. Here, the generator tries to `fool' the discriminator, i.e. make the discriminator predict with a high probability that the generator's output belongs to the input data, while the discriminator is trained to differentiate between the original and the generated samples. The adversarial loss is given in Eqn. \ref{eqn:adloss}.
\begin{equation}
\begin{split}
    L_{adv} = \E_{C_{gt},I_{gray}}[\log\;\mathcal{D}_1 (C_{gt}, I_{gray})] + \\
    \E_{I_{gray}} [\log (1 - \mathcal{D}_1 (C_{pred}, I_{gray}))]
\end{split}
\label{eqn:adloss}
\end{equation}
In Eqn. \ref{eqn:adloss}, $\mathcal{D}_1$ is the discriminator for edge completion GAN. 

The feature matching loss, $L_{FM}$ compares the activation maps in the intermediate layers of the discriminator. This is similar to perceptual loss \cite{johnson2016perceptual} (used in the colour filling model), where activations are compared with those from a pre-trained VGG network. However, a VGG network is not trained to produce the edge maps, so we use the $L_{FM}$ instead of the $L_{perc}$ here. The $L_{FM}$ is defined as :
\begin{equation}
    L_{FM} = \E \left[\sum\limits_{i=1}^L\;\frac{1}{N_i} \left\lVert \mathcal{D}_1^{(i)} \; (C_{gt}) - \mathcal{D}_1^{(i)} \; (C_{pred}) \right\rVert_1\right]
\end{equation}

\begin{figure}
    \centering
    \includegraphics[width = 8.5 cm]{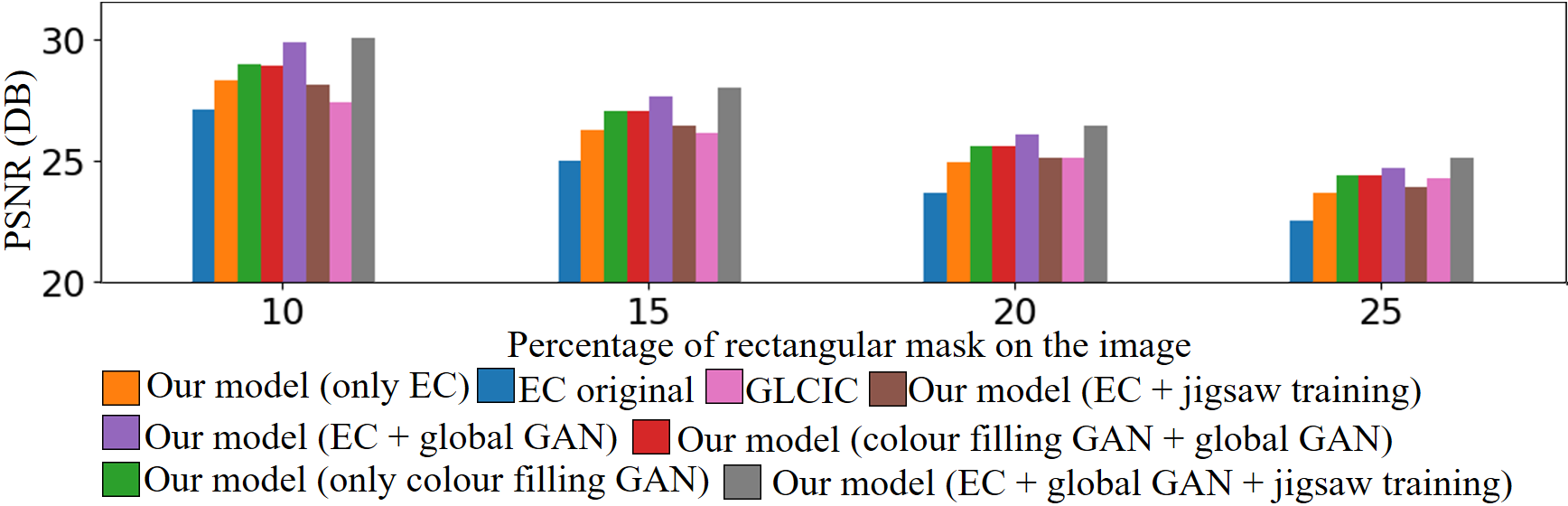}
    \caption{Ablation study on Open Cities AI dataset with rectangular masks to compare the models by removing different components. As expected, the performance is better with all the components.}
    \label{perf_comp_dat}
\end{figure}
Here $L$ is the number of layers in the discriminator. $N_i$ is the number of elements in the $i^{th}$ convolutional layer and $\mathcal{D}_1^{(i)}$ is the activation in the $i^{th}$ layer of the $\mathcal{D}_1$. Applying it to both the $\mathcal{G}_1$ and $\mathcal{D}_1$ gave us better results, as suggested in \cite{nazeri2019edgeconnect}.

The colour filling model is trained on 4 losses. The adversarial loss, L1 loss \cite{zhao2015loss}, style loss \cite{Gatys2016ImageST} and perceptual loss \cite{johnson2016perceptual}. To ensure proper scaling the L1 loss is normalized by the mask size. The adversarial loss is similar to the previous model. The perceptual loss gives the L1 distance between the activations from a few specific layers of a pre-trained network as well as our model. We choose VGG-19 network (pretrained on ImageNet \cite{simonyan2014deep}) since its architecture is similar to ours. 
\begin{equation}
    L_{perc}=\E \left[\sum\limits_{i=1}\;\frac{1}{N_i} \left\lVert \phi_i \; (I_{gt}) - \phi_i \; (I_{pred}) \right\rVert_1\right]    
\end{equation}
where $\phi_i$ represents the activation map of the $i^{th}$ layer of the pre-trained VGG19 network. The style loss returns the differences in covariances of these activation maps by constructing the gram matrix from the activation maps. Given the feature maps of sizes $C_j \times H_j \times W_j$, style loss is computed by
\begin{equation}
    L_{style} = \E_j \left[ \left \lVert G_j^{\phi} \left( \tilde{I}_{pred}\right) - G_j^{\phi} \left(\tilde{I}_{gt}\right) \right \rVert_1 \right]
\end{equation}
where $G_j^{\phi}$ is the gram matrix constructed from activation maps $\phi_j$. The global GAN model is also trained on the weighted sum of the 4 losses as discussed above, namely, $L_{adv}$, $L_{perc}$, $L1$ and $L_{style}$. The final loss for RSINet is given in Eqn. \ref{eqn:loss}. The value of ${\lambda}_{1}$, ${\lambda}_{2}$ and ${\lambda}_{3}$ is empirically fixed to 1 during implementation. 
\begin{equation}
    L_{final} = L_{adv} + {\lambda}_{1}L1 + {\lambda}_{2}L_{perc} + {\lambda}_{3}L_{style}
    \label{eqn:loss}
\end{equation}

\section{Experiments}
The model has a VGG type architecture \cite{simonyan2014deep} (an encoder followed by a decoder). The encoder part consists of 3 convolutional layers each successively halving the image dimensions and doubling the number of channels, followed by 8 residual blocks and finally 3 convolution layers again. This converts the feature representation of the residual blocks back to the image size (the output). A skip connection (consisting of 2 layers) has been added to the second convolutional layer of the network. We have also added 3 gated attention layers in between, for further refinement of the feature maps.

\subsection{Datasets}
We worked primarily with two famous satellite imagery datasets. The first one is the Open Cities AI challenge dataset \cite{naphade20192019}, which consists 500 randomly sampled images size of 1024$\times$1024, divided into 16 images of size 256$\times$256. The second one is the Earth on Canvas dataset \cite{chaudhuri2021zero} which consists of 1400, 256$\times$256 images from 14 classes. It is highly uncorrelated and a good test for our model’s performance. Both the datasets have been divided into 60\%, 20\% and 20\% ratio for train, validation and test sets respectively. 

\subsection{Protocol}
We kept the parameters for the global GAN similar to those of colour filling GAN. Adam optimizer \cite{kingma2014adam} is used for training all the models, with learning rates of 10$^{-3}$ and 10$^{-4}$ for generator and discriminator respectively. We also tried adding jigsaw training \cite{noroozi2016unsupervised}, that refers to splitting the input image into square patches and then randomly shuffling those patches to get a new transformed image. This image would be fed to the model and it would try to reconstruct the original image. For evaluation, peak signal-to-noise ratio (PSNR) is used. For the GLCIC model, its 160$\times$160 output was interpolated to 256$\times$256 while computing the PSNR.
\begin{figure*}[h]
    \centering
    \includegraphics[width=17cm]{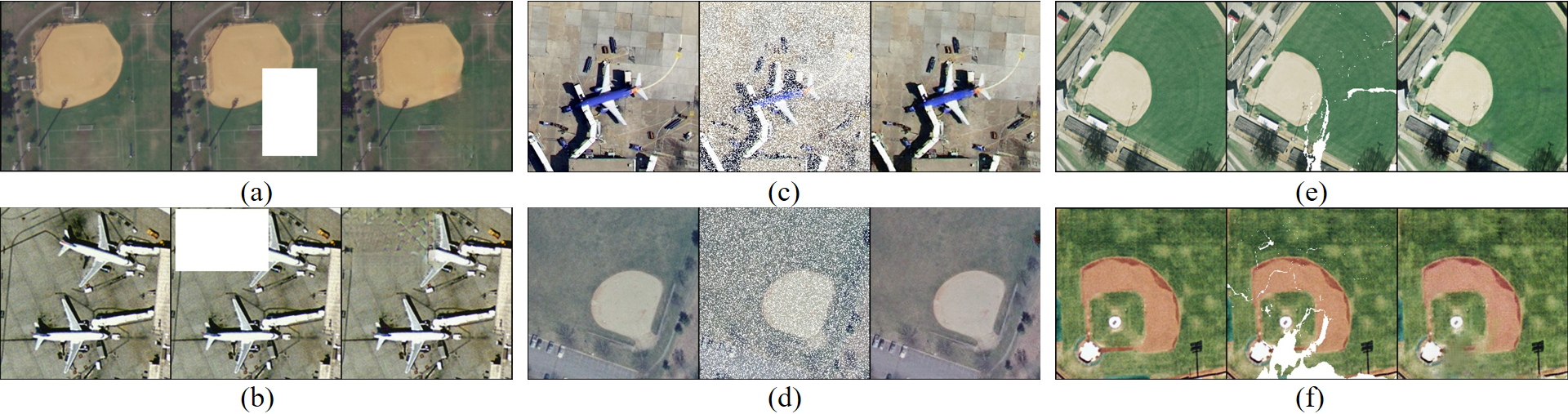}
    \caption{Inpainting results on Earth on Canvas dataset with (a-b) rectangular masks, (c-d) salt-pepper noise, (e-f) irregular masks. For each example, the image on the left is the original image, the image in the centre has been corrupted by a mask while the one in the right is the reconstructed image after inpainting.}
    \label{fig:inpainted}
\end{figure*}

\begin{table}[htbp]
  \centering{\scriptsize
  \caption{\label{tab1} Accuracy analysis for irregular masks on Open Cities AI and Earth on Canvas datasets for different methods}
    \begin{tabular}{|p{3.6cm}|p{1.5cm}|p{1.7cm}|}
    \hline
          & \multicolumn{2}{c|}{PSNR (DB)} \\
    \hline
Model Name & Open Cities AI & Earth on Canvas \\
 \hline
EC original \cite{nazeri2019edgeconnect} & 37.113 & 33.145 \\
EC (our model) & 37.433 & 33.784\\
Only colour filling GAN (our model) & 30.347 & 31.198 \\
PartialConv \cite{liu2018image} & 35.912 & 31.246\\
EC+Global GAN (our model) & \textbf{37.456} & \textbf{34.542} \\
    \hline
    \end{tabular}}
\end{table}

\subsection{Discussions}

As observed in Fig. \ref{perf_comp_dat}, the baseline EC model (i.e. with edge completion GAN and colour filling GAN) along with the global GAN trained on top of it, outperforms the original EC as well as the GLCIC models on the rectangular masks type ablation on both the datasets. The performance for all the models is presented in Table \ref{tab1}. It is visible that our model outperforms all the models for Open Cities AI dataset (highest PSNR of 37.456) and Earth on Canvas dataset (highest PSNR of 34.542).

\subsection{Critical Analysis}
\subsubsection{Ablations}
We primarily dealt with 3 types of ablations; \textit{rectangular masks} wherein a random rectangle of the input image was replaced with white pixels, covering between 5\%-30\% of the image area during training, \textit{salt and pepper masks} where, random image pixels as sampled from a gaussian distribution were whitened, covering anywhere between 5\%-95\% of the image area and \textit{irregular masks}, as shown in \cite{liu2018image}. Furthermore, we also present the visualizations of inpainting for the different masks in Fig. \ref{fig:inpainted} for the three kinds of masks.  The method has shown a good performance in reconstructing the image for \textit{salt and pepper}, and \textit{irregular} masks. However, in case of rectangular mask, in Fig. \ref{fig:inpainted} (b), the model could not construct the \textit{aeroplane} completely. This is because, the method exploits the locally available information (using convolutions) for reconstruction. 


\begin{figure}[!ht]
    \centering
    \includegraphics[width = 8.5cm]{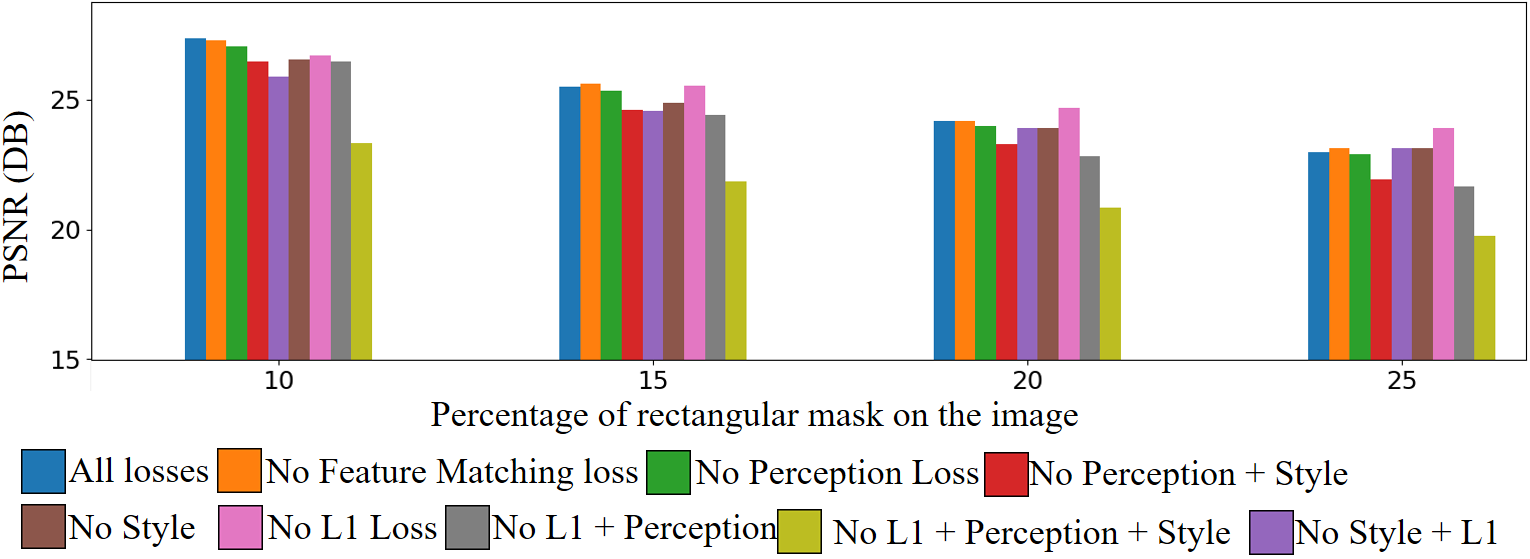}
    \caption{Performance of various loss functions presented on Open Cities AI dataset. It can be observed that relatively better performance is observed when the model is trained on all the losses.}
    \label{loss_perf}
\end{figure}

\subsubsection{Effect of Loss Function}
As can be seen from Fig. \ref{loss_perf}, a mixture of losses works best for inpainting (presented on Open Cities AI). This hybrid loss function consisting of perceptual, style, L1 and adversarial losses for the generator, and feature matching and adversarial losses for the discriminator performs well for this semantically tough image inpainting task. Perceptual and style losses help in improving the quality of the reconstructed image by capturing image semantics and minimising the difference of activations in all the layers of the target and the output images. L1 loss improves the model performance due to its high noise robustness. Feature matching loss used in the discriminator helps to keep the GAN stable and prevent mode collapse. To show the diverse performance, the ablation study is performed for different percentage of rectangular masks. 

\section{Conclusion}
We presented a novel approach to tackle the problem of image inpainting in remote sensing domain. Our approach builds on the existing EdgeConnect model and strengthens it by incorporating attention mechanism and global GAN, that gives realistic results by combining global and local features. Furthermore, we incorporate multiple losses such as perceptual loss and style loss, which further boost our model's performance. We evaluate our model on Open Cities AI and Earth on Canvas datasets, where our approach gives competitive results with respect to previous benchmarks. In future, we would consider more semantically challenging datasets and explore the problem in a multimodal scenario. 

\section*{Acknowledgement}
The authors would like to acknowledgement the IITB-ISRO Grant RD/0120-ISROC00-005.

\bibliographystyle{IEEEbib}
\bibliography{strings,refs}

\end{document}